\documentclass[]{spie}  

\usepackage{amsmath,amsfonts,amssymb}
\usepackage{graphicx}
\usepackage[colorlinks=true, allcolors=blue]{hyperref}
\usepackage{xcolor}
\usepackage{booktabs}

\title{Distance Map Supervised Landmark Localization for MR-TRUS Registration}

\author[1]{Xinrui Song}
\author[1]{Xuanang Xu}
\author[2]{Sheng    Xu}
\author[3]{Baris Turkbey}
\author[2]{Bradford J. Wood}
\author[4]{Thomas Sanford}
\author[1]{Pingkun Yan}

\affil[1]{Department of Biomedical Engineering, Rensselaer Polytechnic Institute, Troy, NY, USA}
\affil[2]{Center for Interventional Oncology, Radiology \& Imaging Sciences, National Institutes of Health, Bethesda, MD, USA}
\affil[3]{Molecular Imaging Program, National Cancer Institute, National Institutes of Health, Bethesda, MD, USA}
\affil[4]{Department of Urology, The State University of New York Upstate Medical University, Syracuse, NY, USA}

\authorinfo{Send correspondence to Pingkun Yan (yanp2@rpi.edu).}

\begin{document} 
\maketitle


\section{Description of purpose}
\label{sec:intro}  

Image-guided interventional procedures often require registering multi-modal images to visualize and analyze complementary information. For example, prostate cancer biopsy benefits from fusing transrectal ultrasound (TRUS) imaging with magnetic resonance (MR) imaging to optimize targeted biopsy. However, cross-modal image registration is a challenging task. This is especially true when the appearance of two image modalities are vastly different. 
Since registration quality is most reliably evaluated with target registration error (TRE), it is sensible to directly make use of the anatomical landmark targets from images. Moreover, while image modalities and thus textures differ, anatomical landmarks are the only information shared across the moving and the fixed images. Sun et al.~\cite{sun2014three} proposed to perform pre-alignment for MR-TRUS registration with manually labeled landmarks on both images. However, such a procedure is still far from automatic due to the requirement of manual input at inference time. Heinrich et al.~\cite{heinrich2022voxelmorph++} made use of a landmark detection method specifically designed for lung computed tomography (CT) registration, which is not generalizable to other tasks.

In this work, we propose to explicitly use the landmarks of prostate to guide the MR-TRUS image registration. We first train a deep neural network to automatically localize a set of meaningful landmarks, and then directly generate the affine registration matrix from the location of these landmarks. For landmark localization, instead of directly training a network to predict the landmark coordinates, we propose to regress a full-resolution distance map of the landmark, which is demonstrated effective in avoiding statistical bias to unsatisfactory performance and thus improving performance. We then use the predicted landmarks to generate the affine transformation matrix, which outperforms the clinicians' manual rigid registration by a significant margin in terms of TRE. 

\section{METHODS}

\begin{figure}
\centering
\includegraphics[width=0.99\textwidth]{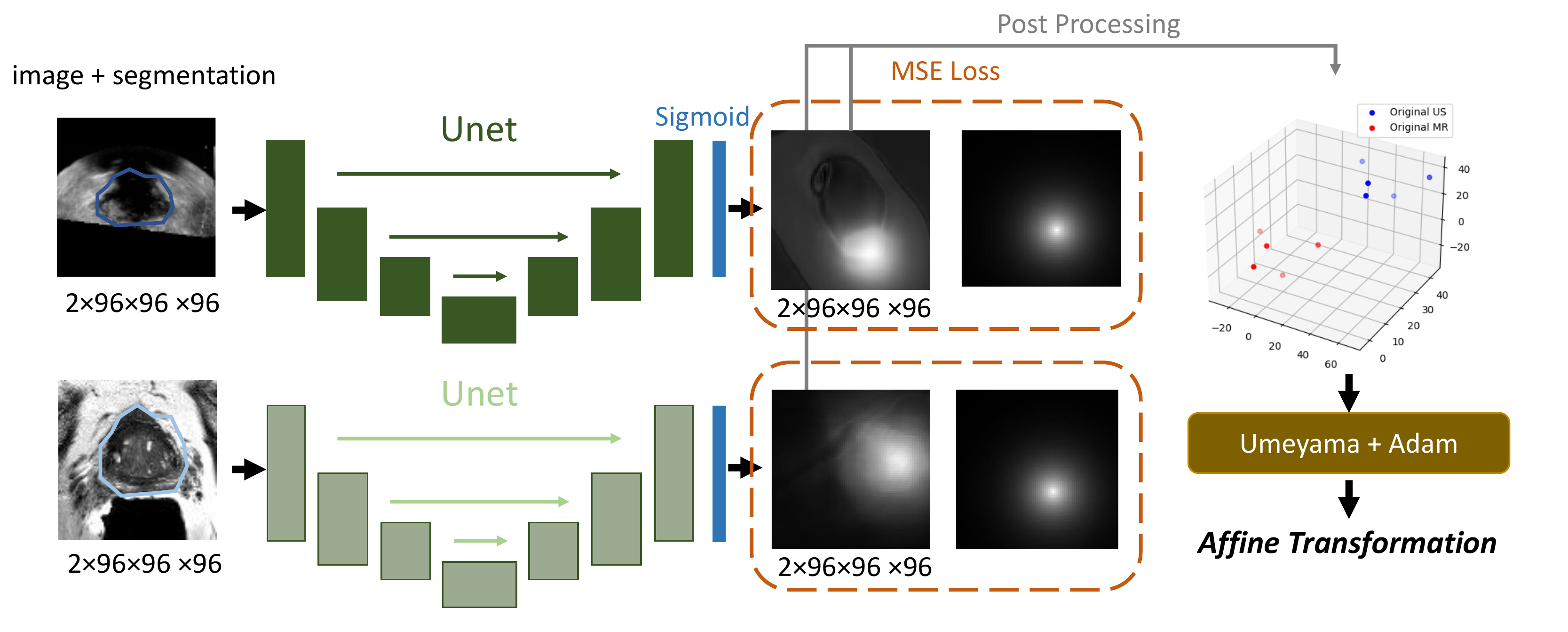}
\caption{Overview of the proposed method} 
\label{fig:flowchart}
\end{figure}

Figure~\ref{fig:flowchart} gives an overview of the proposed method. We first extract the location of selected landmarks from both TRUS and MR images. Then, the two sets of corresponding landmarks are used to calculate an affine registration matrix that aligns the two images. The acquisition of landmarks is explained in Section~\ref{sec:landmark localization}, and the step that generates the affine registration from corresponding points is explained in Section~\ref{sec:affine registration}.

\begin{figure}
\centering
\includegraphics[width=0.9\textwidth]{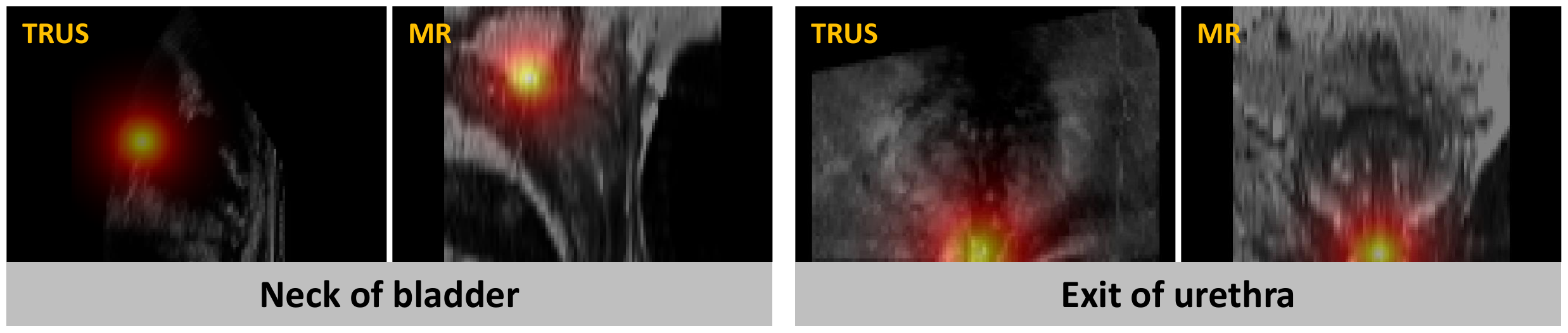}
\caption{Illustration of the two anatomical landmarks' distance map used as label for localization network} 
\label{fig:landmark_label}
\end{figure}

\subsection{Distance map-guided landmark localization}
\label{sec:landmark localization}

Anatomically important and stable landmarks are key to registration, as it is to our method. We referred to previous literature \cite{sun2014three} and consulted clinicians to finally decide on using four landmarks. Two of the landmarks are the extreme points located on the prostate boundary: the right-most and left-most extreme points observed from the axial view. The other two (shown in Figure~\ref{fig:landmark_label}) are the entrance point of the urethra into the prostate at the neck of the bladder, and the point where the urethra exits the prostate at the prostate apex. As the two extreme points of the prostate boundary can be directly extracted from prostate segmentation, which is available for pre-operative MR and can be predicted with established methods for TRUS, we will focus on the localization of the two other anatomical landmarks: the neck of the bladder, and the exit point of the urethra. 

In the proposed method, a network similar to UNet is used to predict a distance map, which is then post-processed to recover the coordinate of the desired landmark. This is different from conventional landmark localization methods \cite{wu2017facial,alansary2019evaluating}, which towards the end of the network, convert the feature map into a feature vector, and then directly regress the landmark coordinates with fully connected layers. Our experiments show that such an approach is prone to statistical biases when applied to 3D volumes, and could collapse to always outputting the mean value of the training samples. 
To overcome this issue, we designed a distance map, paired with UNet, to guide the training of the network. The skip connections and encoder-decoder structures of UNet enforce the model to make more explicit use of the image contents. More importantly, the prediction of the network is a full-resolution map that corresponds to the input image, resembling the human labeling process of drawing onto the image, as opposed to the abstract procedure of directly regressing the coordinates from the fully connected layers. We also added a sigmoid layer at the end of the UNet to normalize the output to a range of [0, 1].  

\begin{equation}
\label{eqn:gt distance map}
    map_{label}= exp(-10\times\frac{M}{max(M))}), M = distance \: map
\end{equation}

The distance map is first calculated with Maurer's algorithm~\cite{maurer2003linear}. We then normalize the distance map into the label map according to Equation~\ref{eqn:gt distance map}, so that the actual location of the landmark will have a value of one on the label map, and the surrounding voxels gradually fade to zero in all directions. We supervise the network training with MSE loss calculated between the label map and the predicted map. 


\subsection{Affine registration}
\label{sec:affine registration}

The Kabsch–Umeyama algorithm \cite{umeyama1991least} is capable of finding the transformation parameters, including translation, rotation, and uniform scaling that give the least mean square error from two sets of corresponding points. In the proposed method, we first use the predicted landmarks to determine the optimal transformation parameters through the Umeyama algorithm, then apply this transformation to the ground truth landmark points to calculate the TRE for evaluation. 

\begin{equation}
\label{eqn:optimization}
argmin_{\theta }\left \|  P_{TRUS} - T_{\theta}( P_{MR}) \right \|_{L2}
\end{equation}

Although the Kabsch–Umeyama algorithm is an established method, it is still limited to uniform scaling, which is far from real-world deformations. In this paper, we propose to use the results of the Umeyama algorithm as a starting point for an optimization-based method that allows nonuniform scaling. We first decompose the matrix produced by the Umeyama algorithm into nine transformation parameters, then used the Adam optimizer \cite{kingma2014adam} to perform gradient-descent optimization on the transformation parameters. We use the euclidean distance between the two sets of points as the loss function. This method enables fully-affine registration and leads to better results as shown in Section~\ref{sec:results}. The objective function for optimization is shown in Equation~\ref{eqn:optimization}.

\section{RESULTS}
\label{sec:results}

\subsection{Implementation details}
\label{sec:implementation}
We divide our dataset to a training set of 21 cases, validation set of 3 cases, and test set of 5 cases. All images are resampled to $96\times96\times96$ with isotropic spacing for network input. For Adam gradient descent optimization, we consistently used $1\times10^{4}$ iterations with a step size of $1\times10^{-5}$. 

\subsection{Experiment results}
\label{sec:Experiment results}
In table \ref{table:4TRE} we present the result of the proposed registration method in comparison to no registration and manual registration by clinicians. The first row of the table shows that if the two sets of landmarks are known, the proposed method of using gradient descent optimization to finetune the result of the Umeyama algorithm achieves better results. Such improvement results from allowing non-uniform scaling, which isn't accessible through the Umeyama algorithm. 

\begin{table}
\caption{\label{table:4TRE}Registration performance of different methods. Measurements are TRE values in mm.}
\centering
\begin{tabular}{c|c|c|c}
\toprule
landmarks    & TRUS segmentation & Umeyama & \textbf{Umeyama+Adam} \\ \midrule 
manual   & manual& 4.02$\pm$0.82 & 3.71$\pm$0.77\\
predicted    & manual  &  4.41$\pm$0.81  & 4.07$\pm$0.67\\
predicted   & predicted & 4.93$\pm$1.18 &4.68$\pm$1.39\\
\midrule 
Before registration & \multicolumn{3}{c}{7.53$\pm$2.11} \\
Manual rigid registration & \multicolumn{3}{c}{6.31$\pm$1.28} \\
\bottomrule
\end{tabular}
\end{table}

The third row of Table \ref{table:4TRE} shows the result of using predicted TRUS segmentation to extract the extreme points, and as an auxiliary input to the landmark localization network. The second row shows a scenario in which the ground truth TRUS segmentation is available, and only the landmarks are predicted through neural networks. Although the results on the third row are generally larger in error than the second row, it is still significantly lower in comparison to the TRE before registration, or the result of manual registration ($p\: value < 0.05$ through paired T-test). Additionally, the TRUS segmentation network is trained on a small dataset. Given the better accessibility of TRUS prostate segmentation in comparison to landmark segmentations, it is perfectly feasible to improve the result of the third row with a TRUS segmentation network trained on a larger dataset. The overall performance of the pipeline can also be improved with B-spline alignments of landmarks on top of affine transformations. We will explore these ideas in the future.

\begin{table}
\caption{\label{table:landmark localization}Landmark localization performance comparison. Measurements are TRE values in mm.}
\centering
\begin{tabular}{c|c|c}
\toprule
Methods    & Neck of bladder & Urethra exit point \\ \midrule 
ResNet + coordinate label \cite{wu2017facial,alansary2019evaluating}   & 6.21 & 6.24 \\
\textbf{UNet + distance map label}    & $\mathbf{3.12}$  &  $\mathbf{3.377}$ \\
\bottomrule
\end{tabular}
\end{table}

In Table \ref{table:landmark localization}, we compare the performance of the proposed distance map-guided landmark localization method with the conventional approach of directly predicting the landmark coordinate. Our experiments show that for 3D medical images, such as MR and TRUS as in our dataset, the proposed method outperforms its conventional counterpart. 
In Table \ref{table:cyst TRE}, we compare the TRE calculated with landmarks not included in the process of affine registration prediction. In this case, we are using the center points of cysts as the evaluation landmark. This experiment demonstrates the generalizability of the proposed registration pipeline. Even if a landmark is left outside the process of registration prediction, the registration based on other landmarks ultimately leads to a universally better landmark alignment. The proposed method in this table refers to the same method from the third row of Table \ref{table:4TRE}, with both the landmarks and the TRUS segmentation predicted through neural networks. The registered TRE outperforms manual registration and is on par with that calculated from manually labeled landmarks. 

\begin{table}
\caption{\label{table:cyst TRE}Registration error of landmarks not used for registration. Measurements are TRE values in mm.}
\centering
\begin{tabular}{c|c|c|c|c}
\toprule
Methods    & No registration & Manual rigid registration & \textbf{Proposed} & \textbf{Manual landmarks + Adam} \\ \midrule 
    TRE of cyst   & 4.92$\pm$1.12 & 4.40$\pm$0.59& $\mathbf{2.42\pm1.19}$ & $\mathbf{2.34\pm1.01}$ \\
\bottomrule
\end{tabular}
\end{table}

\section{New or breakthrough work to be presented}
\label{sec:breakthrough}
The contributions of this study are three-fold.
1) We propose a new landmark-guided pipeline for automatic MR-TRUS registration. 
2) We propose a new landmark localization method that is more suitable for 3D medical images.
3) We use an optimization method to allow non-uniform scaling in landmark-guided affine registration, thereby improving registration quality. 

\section{CONCLUSION}

In this paper, we present a novel pipeline to guide multi-modal registration, which utilizes a novel landmark localization method that outperforms conventional methods. The proposed registration pipeline itself also shows great potential to be improved even further in future studies. 

\textbf{The authors confirm that this work has not been submitted for publication elsewhere.}

 

\bibliography{main} 
\bibliographystyle{spiebib} 

\end{document}